\documentclass{article}
\usepackage{interspeech2005,amssymb,amsmath,epsfig,tipa}
\usepackage[usenames]{color}
\ninept
\def\reg{{\rm\ooalign{\hfil
     \raise.07ex\hbox{\scriptsize R}\hfil\crcr\mathhexbox20D}}}

\title{Using phonetic constraints in acoustic-to-articulatory inversion} 

\makeatletter
\def\name#1{\gdef\@name{#1\\}}
\makeatother
\name{{\em Blaise Potard, Yves Laprie}}

\address{LORIA - Campus Scientifique  \\
BP 239 - 54506 Vandoeuvre-les-Nancy Cedex \\
{\small \tt potard@loria.fr}
}
\begin{document}
\maketitle
\begin{abstract}
The goal of this work is to recover articulatory information
from the speech signal by acoustic-to-articulatory inversion. One
of the main difficulties with inversion is that the problem
is under-determined and inversion methods generally offer no guarantee
on the phonetical realism of the inverse solutions. A way 
to adress this issue is to use additional phonetic constraints. 

Knowledge of
the phonetic caracteristics of  French vowels enable the derivation of
reasonable articulatory domains in the space of Maeda  parameters:
given the formants frequencies (F1,F2,F3) of a speech sample, and thus
the vowel identity, an ``ideal'' articulatory domain can be derived.
The space of formants frequencies is partitioned into vowels, using either speaker-specific
data or generic information on formants. Then, to each articulatory vector
can be associated a phonetic score varying with the distance to
the ``ideal domain'' associated with the corresponding vowel. 

Inversion experiments were conducted on isolated vowels and
vowel-to-vowel transitions. Articulatory parameters were compared with
those obtained without using these constraints and
those measured from X-ray data.
\end{abstract}

\section{Introduction}
Atal and his colleagues\cite{Atal78} have shown that an infinity of
area functions can give exactly the same 3-tuple of formants. One of
the challenges in acoustic-to-articulatory inversion is thus to add
constraints which reduce the number of inverse solutions without
eliminating relevant solutions. One common approach is to use an
articulatory model that generates only relevant vocal tract
shapes. These 2D or 3D models are generally derived from medical
images acquired for one subject by applying some factor analysis
technique. Even if an articulatory model substantially reduces the
range of possible vocal tract shapes there still exists a very large
number of inverse solutions for each 3-tuple of formants.

Actually, it turns out that the articulatory variability is one of the
essential characteristics of speech production. The articulators of
speech have large compensation capacities that enable the production
of one sound one after the other even if its intrinsic articulatory
characteristics are very far from those of the other. Despite this
large variability there exist a number of expected articulatory
invariants. The aim of the work reported in this paper is to exploit
standard phonetic knowledge to express these articulatory invariants
in the form of constraints imposed to articulatory parameters.

Other classes of constrains have been investigated. Physiological
constraints, for instance, give ranges of possible articulatory
parameters and/or constraints about the maximal acceleration or jerk
(third derivative of position) acceptable for speech. However, most of
these constraints require the knowledge of parameters that cannot be
easily accessible. The main advantage of phonetic constraints is that
they can be easily expressed and that they present a great robustness
with respect to speaker variability. 

At first we
describe the phonetic constraints and their implementation in our
acoustic-to-articulatory framework\cite{laprie03f}, which uses an
articulary table (or codebook), generated using Maeda's articulatory
model\cite{maeda79}. Then we
evaluate them in the case of isolated vowels to investigate their
effects in terms of place and degree of constriction, and in the case
of speech utterances for which the articulatory parameters are known.

\section{Phonetic features as articulatory constraints}


The main idea behind the use of phonetic constraints is the assumption
that  each phoneme has  invariant articulatory features, like a strong
protrusion for the french \textipa{/y/}, for instance. In the case of vowels,
which present slow time varying acoustic structures in comparison to
other phonemes as stop consonants, these features can be easily
translated into constraints on the articulatory parameters.

\subsection{Phonetic constraints for vowels}

In the particular case of vowels, four types of constraint can be
defined : the mouth opening, the protrusion of the lips, the lip
stretching, and the position of the tongue dorsum. The relevance of
each constraint depends on the vowel considered. As mentioned in the
introduction there exists a strong inter-speaker variability. We thus
designed numerical, rather than boolean, constraints that return a
phonetic relevancy from the knowledge of formants. 

Tab. \ref{tablevowels} summarizes our classification for the 10
non-nasals French vowels. $D$ stands for ``tongue dorsum position'', $O$ for ``mouth
opening'', $S$ for ``lip stretching'', and $P$ for ``lip protrusion''. The
convention we use for classification is straightforward: the higher
the number, the higher the value associated with the given
constraint. For example, a constraint $O_1$ means that the mouth has a
small opening, a value of $O_4$ means a very big opening. These data
are average values of the way native French speakers articulate
vowels, and thus may be  different from the way a particular
speaker articulates sounds of French. Note that for the main place of
articulation of vowels, corresponding to $D$ in the case of vowels,
the range of possible values  is a sub-domain of the values acceptable
for consonants (from 0 for /p,b,m/ to 9 for /\textinvscr,
\textturnr/). This explains why $D$ ranges only between 6 and 8 for
vowels.

\begin{table} [t,h]
\caption{\label{tablevowels} {\it French vowels classification.}}
\vspace{2mm}
\centerline{
\begin{tabular}{|c|c|c|c|c|}
        \hline
        Vowel & D & O & S & P \\
        \hline
        \textipa{i} & D6 & 01 & S4 & P1 \\
        \textipa{e} & D6 & 02 & S3 & P1 \\
        \textipa{E} & D6 & 03 & S2 & P1 \\
        \textipa{a} & D7 & 04 & S1 & P1 \\
        \hline
        \textipa{y} & D6 & 01 & S1 & P4 \\
        \textipa{\o} & D6 & 02 & S1 & P3 \\
        \textipa{\oe} & D6 & 03 & S1 & P2 \\
        \hline
        \textipa{u} & D8 & 01 & S1 & P4 \\
        \textipa{o} & D8 & 02 & S1 & P3 \\
        \textipa{O} & D8 & 03 & S1 & P2 \\
        \hline
        \end{tabular}}
\end{table}

\subsection{Transposing phonetic constraints in the articulatory model}

In most articulatory models, transposing simple phonetic features into
parameters of the model can be quite complex. In our case, we use
Maeda's model\cite{maeda79}, in which the parameters can be easily
interpretable from a phonetic point of view. Consequently, expressing
phonetic constraints in terms of articulatory parameters is straightforward:
lip protrusion and tongue dorsum position are already
parameters of the model, and the mouth opening is a linear combination
of two parameters (jaw position, and intrinsic
lip opening). 

Actually, this constraint also uses the tongue position
in order to take into account compensatory effects described in
\cite{Maeda90book}: Maeda observed that for non-rounded vowels (/i/,
/a/, /e/), the
tongue position and the jaw opening had parallel effects on the
acoustic image, and therefore were mutually compensating. He also
observed that this compensatory effect was indeed used by
his test subjects. Furthermore, it appeared that the direction of
compensation did not depend on the vowel pronounced: there was a
linear correlation \[Tp + \alpha Jw = \textrm{ Constant}\], where $Tp$ is the tongue position, $Jw$ the jaw position, and the $\alpha$ the linearity coefficient
that is the same for both /a/ and /i/. The other vowels were not studied
because there were not enough occurrences of them in the X-ray
database. Maeda observed this compensation in both his subjects (but the
coefficients of correlation were of course different). The coefficient
we used for PB was the one Maeda found experimentally on X-ray data, which was
approximately equal to $0.66$. This compensatory effect allowed Maeda
to explain most of the articulatory variability for /a/ and /i/.


\subsection{Acoustic space partitionning}


For each phoneme, we have to define an acoustic domain where the
phonetic constraints are considered to be valid, that is, a domain
where we are likely to observe articulatory configurations which
respect the given constraints. We could compute these domains directly
from the articulatory model, by synthesizing the domains of the phonetic
constraints: in future works, we may use self-organising maps
like Kohonen's. But currently, we  use  simple models,
centered on the average vowels formant frequencies of French
speakers. 

\begin{figure}[t]
\centerline{\epsfig{figure=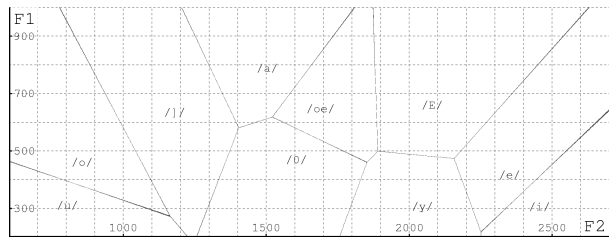,width=80mm}}
\caption{{\it Voronoi diagram model.}}  
\label{voronoi}
\end{figure}

\begin{figure}[t]
\centerline{\epsfig{figure=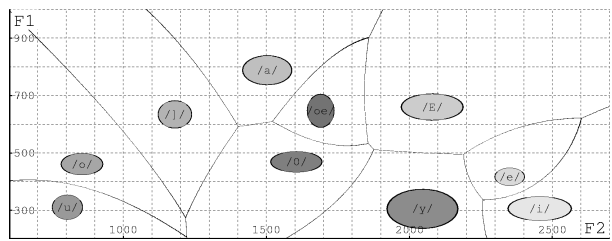,width=80mm}}
\caption{{\it Ponderated Voronoi diagram model.}}  
\label{ponderated}
\end{figure}

Currently, our model works on the 3-D space of the  first three formant
frequencies. We tested different models for the
partitionning of the acoustic space : Voronoi diagram around the vowels
(cf. Fig. \ref{voronoi}); Voronoi diagram weighted by the standard
deviation of each formant frequencies (cf. Fig. \ref{ponderated}.

\subsection{Phonetic scoring}

Now that we have partionned the acoustic space, we still have to
explain how a phonetic score can be associated to each inverse
solution: basically, a given acoustic vector is attached to an ``ideal
articulatory domain'', as defined by the constraints in
Tab.~\ref{tablevowels}, corresponding to the region of the acoustic space it
belongs to. Then each inverse solution $V$ corresponding to this 3-tuple
can be given a ``phonetic score'', according to the distance of the
articulatory vector to the ``ideal domain''. A simple way to do that
would be to compute the norm of the vector defined by the point and
its orthogonal projection onto the domain. Actually, we compute a score
relative to each type of constraint: tongue dorsum, mouth opening,
lip stretching and protrusion. 

The computation of the score depends on two values: the
target value of the constraint considered $\theta(v,t)$, where $v$ is the vowel, and $t$ is the
type of constraint considered, and a margin $\sigma(v,t)$, which defines
a validity interval $I(v,t) = [\theta(v,t) - \sigma(v,t); \theta(v,t) +
\sigma(v,t)]$. If the value of the constraint for $V$ is within
$I(v,t)$, then it gets a perfect score ($1$) for that type of
constraint. Otherwise, it gets a positive score less than 1 which exponentially decreases from 1 according  to the
distance to $I(v,t)$. The overall phonetic score is simply a linear combination of the
4 types of constraints,  to get scores within the interval
$[0;1]$ (1 being the best score). In our current model, all constraints have equal weight, except for
the lip stretching which has a null weight, because Maeda's model
cannot account for lip stretching, since it was designed using X-ray
images of sagittal profiles of the vocal tract.

\section{Experiments}

We conducted inversion experiments on the original data Maeda used for
his model. It consisted in a corpus of 10 sentences for a total time
of about 20 seconds of X-ray cineradiography. Cardinal vowels and some
VV sequences were selected in the speech signal, the first three
formants frequencies were manually extracted. We built a high precision
codebook adapted to Maeda's speaker. Although we studied the original
speaker used to build the articulatory model, we still had to adapt the
model to improve the acoustic faithfulness\cite{laprie03f} because the
geometrical calibration of the X-ray acquisition is not known
precisely.

Despite this adaptation it must be kept in mind that the articulatory model
together with the acoustic simulation are not capable of generating formant
frequencies that have been measured from the original speech signal.  Even
by using articulatory parameters measured from X-ray images and the best
geometrical adaptation the average error on F1 is still 54\,Hz. This non
negligible discrepancy is explained by the approximation  of the recovery
of the 3D information (corresponding to the area function) from the 2D
information (corresponding to the sagittal profile of the vocal tract)
provided by the  articulatory model. This approximation, based on the
method proposed by Heinz and Stevens\cite{heinz1965}, is unable to render
the area everywhere from the glottis to lips precisely. In addition,
physical constants involved in the acoustic simulation probably introduce
a slight error. 
In conclusion, despite this favourable situation (the speech signal to
be inverted has been pronounced by the speaker whose X-ray data have
been processed to derive the articulatory model) the inversion is non trivial
and cannot precisely recover the original articulatory trajectories.


\subsection{Codebook caracteristics}

Tab. \ref{caracod} summarises the characteristics
of the codebook used for inversion. The first line gives the number of unique
articulary vectors which acoustic image was calculated during the
codebook construction. The second line gives the number of linear
hypecubes which were kept in the codebook. The third line gives the
total\footnote{the actual number of unique articulatory vectors is
lower than this number, which is simply the number of hypercubes
multipled by the number of vertexes in an hypercube, that is, $2^7 = 128$.}
number of vertexes of the forementioned hypercubes. The fourth line gives
the percentage of the total volume of hypercubes of the codebook over
the whole articulatory space explored. The fifth line gives the
maximum (over the first three formant frequencies) average
absolute error of the formant frequencies  
linearly interpolated from the codebook data over the formants
computed using the articulatory model. The acoustic precision used in
the codebook construction for
the linearity test was 0.3 bark on each formant frequency.

\begin{table} [t,h]
\caption{\label{caracod} {\it Codebook characteristics.}}
\vspace{2mm}
\centerline{
\begin{tabular}{|l|l|}
\hline
Number of points sampled & 607,422,368 \\
\hline  
Number of hypercubes & 1,071,353 \\
\hline
Number of vertexes & 137,133,184 \\
\hline
Articulatory space kept & 32.9 \% \\
\hline
Average acoustic precision & 8.3 Hz \\
\hline
\end{tabular}}
\end{table}

\subsection{Checking the model consistency}

As the phonetic constraints, as well as the acoustic space
partitionning, are independant of the speaker in our current model, we beforehand 
checked that the acoustic domains correspond to the images of the
phonetic constraints domains. For each vowel, we plotted the acoustic
images of articulatory vectors that had perfect phonetic scores, and
we could observe that for each vowel, the acoutic domain was included in the overall
acoustic image of the corresponding ``ideal'' articulatory
domain. We also computed a new partition of the acoutic space by
attributing each point of the acoutic space to the vowel that had
most images in its neighborhood (each vowel had the same number of
synthesized articulatory points, randomly chosen in the ideal
domain). The resulting F1/F2 graph was very 
close to our acoustic models.


\subsection{Inversion of isolated vowels}

\begin{figure}[t]
\centerline{\epsfig{figure=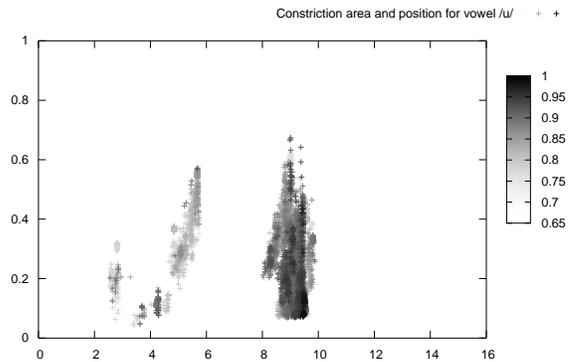,width=60mm, angle=-90}}
\caption{{\it Phonetic scoring of the inverse solutions for
/u/}}  
\label{u_constrict}
\end{figure}

Vowels
/\textipa{a}/,/\textipa{i}/,/\textipa{u}/,/\textipa{e}/,
/\textipa{o}/, were inverted using the phonetic constraints. The
inverted points are each given a phonetic score varying with their
distance from the ``ideal domain''. Fig. \ref{u_constrict}
represents the area at the maximum constriction (in cm$^2$) as a function of
its position (in cm, starting from the glottis) for each
inverse solution found. The gray level of each point is a function of its
phonetic score, darker points have a higher score. 
Although constraints are applied on articulatory parameters, they give rise
to a consistent overall effect, i.e. they enhance the emergence of well
located regions in the plane spanned by  the place of maximal constriction and the constriction area, and weakens some secondary places of articulation. These regions are furthermore more consistent with the articulatory data of Wood\cite{wood79}.
The second  observation is that these phonetic constraints penalize
vocal tract shapes with large constriction areas. This aspect is 
important because the acoustic properties of vocal tract shapes are
not very sensitive to a general and uniform area increasement. This
thus enables this kind of unrealistic vocal tract shapes to be
penalized.

\subsection{Inversion of VV sequences}

We extracted several VV sequences from the sentences uttered by PB:
/\textipa{ui}/, /\textipa{yi}/, /\textipa{ie}/.

Since
the audio signal was quite noisy, we had to extract  formants
 by hand. After that, the sequences were inverted using
different kinds of constraint. Here, we present the results for
the sequence /\textipa{yi}/. For all the figures the time unit is {\it
ms} and the articulatory parameters are given in standard deviation
with respect to the neutral position. 

\begin{figure}[t]
\centerline{\epsfig{figure=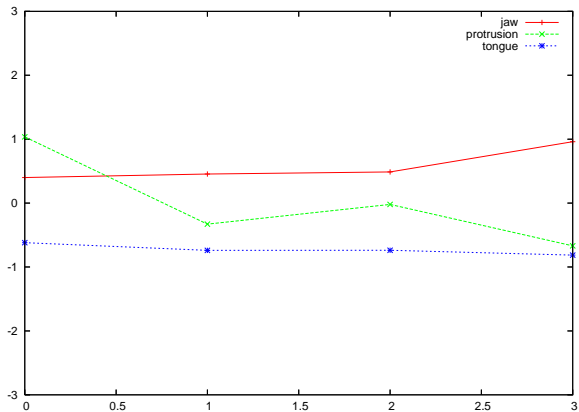, angle=270,width=80mm}}
\caption{{\it Measured articulatory parameters}}  
\label{params}
\end{figure}

Fig. \ref{params} represents the 3 main parameters
(jaw, tongue position, lip protrusion) as measured on the
X-ray images.

\begin{figure}[t]
\centerline{\epsfig{figure=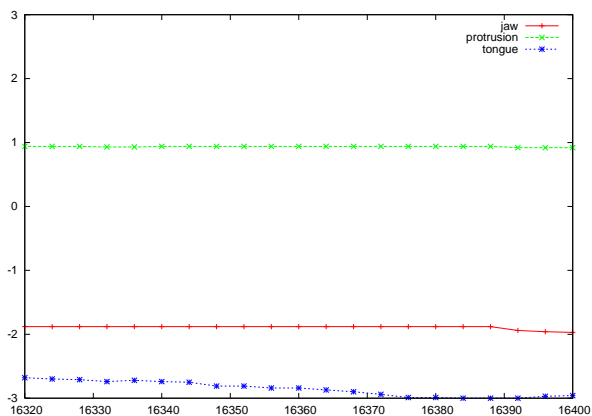, angle=270,width=80mm}}
\caption{{\it Inversion with biodynamic constraints only}}  
\label{dynamic}
\end{figure}

Fig. \ref{dynamic} is the inverted sequence using only
biodynamic constraints on the articulatory parameters: that is, the
``overall velocity'' of  articulators is minimized. Although the
inverse solution has a very good acoustic precision, it is very different
from the observed solution, and it is not phonetically realistic. Not
surprinsingly,  the minimization of the overall velocity gives rise to
quasi-straight transitions.

\begin{figure}[t]
\centerline{\epsfig{figure=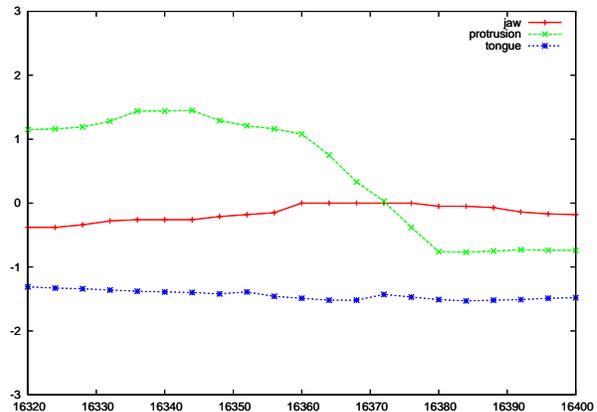, angle=270, width=80mm}}
\caption{{\it Inversion with phonetic and biodynamic constraints}}  
\label{phonetic}
\end{figure}

Fig. \ref{phonetic} is the inverted sequence, using both
biodynamic and phonetic constraints, with equal weights. It should be
noted that the original trajectories are sampled at a lower rate
(50\,Hz) than the inverse trajectories. This time, the solution is
much more realist. The overall articulatory movements have been
recovered properly even if absolute values of the articulatory
parameters are not equal to the original ones. As mentioned above this
is due to the acoustic mismatch between the articulatory acoustic
simulation and the human process of speech production. 

This experiment shows that very general constraints, derived from
standard phonetic knowledge, enable the recovery of realistic
articulatory trajectories. 
The impact of these phonetic constraints is all the more sensitive
since our inversion method exploits a quasi exhaustive description of
the articulatory space.


\section{Conclusion and perspectives}
The  under determination of the acoustic-to-articulatory problem has
given rise to several directions of research in order to incorporate
constraints that can compensate for the lack of data. However, most of
the constraints envisaged (see \cite{sorokin2000} for instance)
require the knowledge of numerical constants difficult to be
estimated. In comparison with these constraints phonetic constraints
present two advantages. Firstly, they do not involve numerous
numerical parameters, which is a key point. Secondly, they are very
general, speaker independent and have been extensively validated since
they derive from standard phonetic knowledge.
Furthermore, these phonetic constraints could be easily coupled with
constraints derived from the observation of the speaker's face.

\begin{center}
  \textbf{Acknowledgments}
\end{center}
We would like to thank Dr. Shinji Maeda fruitful discussions and for making his articulatory
model and data available.

\bibliographystyle{IEEEtran}

\begin{thebibliography}{1}

\bibitem[1]{Atal78}
B.~S. Atal, J.~J. Chang, M.~V. Mathews, and J.~W. Tukey, ``Inversion of
  articulatory-to-acoustic transformation in the vocal tract by a
  computer-sorting technique,'' \emph{JASA}, vol.~63, no.~5, pp. 1535--1555,
  May 1978.

\bibitem[2]{laprie03f}
Y.~Laprie, S.~Ouni, B.~Potard, and S.~Maeda, ``{Inversion experiments based on
  a descriptive articulatory model},'' in \emph{{6th International Seminar on
  Speech Production}}, Sydney, Autralia, Dec. 2003.

\bibitem[3]{maeda79}
S.~Maeda, ``Un mod{\`e}le articulatoire de la langue avec des composantes
  lin{\'e}aires,'' in \emph{Actes 10{\`e}mes Journ{\'e}es d'Etude sur la
  Parole}, Grenoble, Mai 1979, pp. 152--162.

\bibitem[4]{Maeda90book}
------, ``Compensatory articulation during speech: Evidence from the analysis
  and synthesis of vocal-tract shapes using an articulatory model,'' in
  \emph{Speech Production and Speech Modelling}, W.~J. Hardcastle and
  A.~Marschal, Eds.\hskip 1em plus 0.5em minus 0.4em\relax Kluwer Academic
  Publishers, 1990.

\bibitem[5]{heinz1965}
J.~M. Heinz and K.~N. Stevens, ``On the relations between lateral
  cineradiographs, area functions and acoustic spectra of speech,'' in
  \emph{Proceedings of the 5th International Congress on Acoustics}, 1965, p.
  A44.

\bibitem[6]{wood79}
S.~Wood, ``A radiographic analysis of constriction locations for vowels,''
  \emph{Journal of Phonetics}, vol.~7, pp. 25--43, 1979.

\bibitem[7]{sorokin2000}
V.~Sorokin, A.~Leonov, and A.~Trushkin, ``Estimation of stability and accuracy
  of inverse problem solution for the vocal tract,'' \emph{Speech
  Communication}, vol.~30, pp. 55--74, 2000.

\end{thebibliography}

\end{document}